\documentclass[letterpaper, 10 pt, conference]{ieeeconf}  % Comment this line out if you need a4paper

\IEEEoverridecommandlockouts                              % This command is only needed if 
\usepackage{graphics} % for pdf, bitmapped graphics files
\usepackage{subcaption}
\usepackage{epsfig} % for postscript graphics files
\usepackage{mathptmx} % assumes new font selection scheme installed
\usepackage{times} % assumes new font selection scheme installed
\usepackage{amsmath} % assumes amsmath package installed
\usepackage{amssymb}  % assumes amsmath package installed
\usepackage{booktabs}
\usepackage{multirow}
\usepackage{multicol}
\usepackage{hyperref}
\usepackage[table,x11names]{xcolor}

\definecolor{lightblue}{RGB}{187,204,238}
\definecolor{lightgreen}{RGB}{204,221,170}
\definecolor{lightred}{RGB}{255,204,204}

\title{\LARGE \bf
Quantifying and Modeling Driving Styles in Trajectory Forecasting
}

\author{Laura Zheng*, Hamidreza Yaghoubi Araghi*, Tony Wu,\\Sandeep Thalapanane, Tianyi Zhou, and Ming C. Lin \\
\thanks{*Equal Contribution. The authors are with the Department of Computer Science, University of Maryland at College Park, MD, U.S.A. 
    E-mails: \{lyzheng,yaghoubi,tonywu,sandeept,tianyi,lin\}@umd.edu} 
\href{https://gamma.umd.edu/traj_style_analysis/}{\texttt{gamma.umd.edu/traj\_style\_analysis}}
}

\begin{document}

\maketitle
\thispagestyle{empty}
\pagestyle{empty}

\begin{abstract}

Trajectory forecasting has become a popular deep learning task due to its relevance for scenario simulation for autonomous driving. 
Specifically, trajectory forecasting predicts the trajectory of a short-horizon future for specific human drivers in a particular traffic scenario. 
Robust and accurate future predictions can enable autonomous driving planners to optimize for low-risk and predictable outcomes for human drivers around them. 
Although some work has been done to model driving style in planning and personalized autonomous polices, a gap exists in explicitly modeling human driving styles for trajectory forecasting of human behavior. 
Human driving style is most certainly a correlating factor to decision making, especially in edge-case scenarios where risk is nontrivial, as justified by the large amount of traffic psychology literature on risky driving. 
So far, the current real-world datasets for trajectory forecasting lack insight on the variety of represented driving styles. While the datasets may represent real-world distributions of driving styles, we posit that fringe driving style types may also be correlated with edge-case safety scenarios. 
In this work, we conduct analyses on existing real-world trajectory datasets for driving and dissect these works from the lens of driving styles, which is often intangible and non-standardized.

\end{abstract}

\section{Introduction}

% Explain the special domain of traffic systems
Vehicle traffic systems are highly constrained, yet complex. 
In terms of constraints, vehicle kinematics strictly defines what outcome trajectories are physically possible, and traffic laws are a high-level guideline for how traffic should interact at junctions. 
While traffic laws provide general constraints on driving behavior, the space of human decision making within reasonable bounds is high-dimensional and difficult to model. 
Since driving is a time-dependent nonlinear system, modeling human decision making is both unintuitive and challenging. Due to this, trajectory prediction with deep learning has become a popular approach to modeling probable traffic outcomes.

% Explain issues with current approaches
Current state-of-the-art deep learning approaches for trajectory prediction learn traffic, vehicle, and interaction dynamics jointly with deep neural networks. 
Human decision making, vehicle constraints, and traffic laws are often learned implicitly through sheer generalization capabilities of modern deep learning architectures, accompanied with large and robust real-world datasets. 
While these methods often work well in real-world benchmarks at scale, we emphasize that a key gap in these benchmarks is precisely that they are sampled from safe driving in the real world and often represent the most general cases. 
In reality, the most difficult scenarios to predict are fringe-case scenarios, perhaps with an out-of-distribution set of drivers, or with the presence of a stressor in the scenario (such as a jaywalker or a deer crossing). 

\begin{figure}
    \centering
    \includegraphics[width=0.9\linewidth]{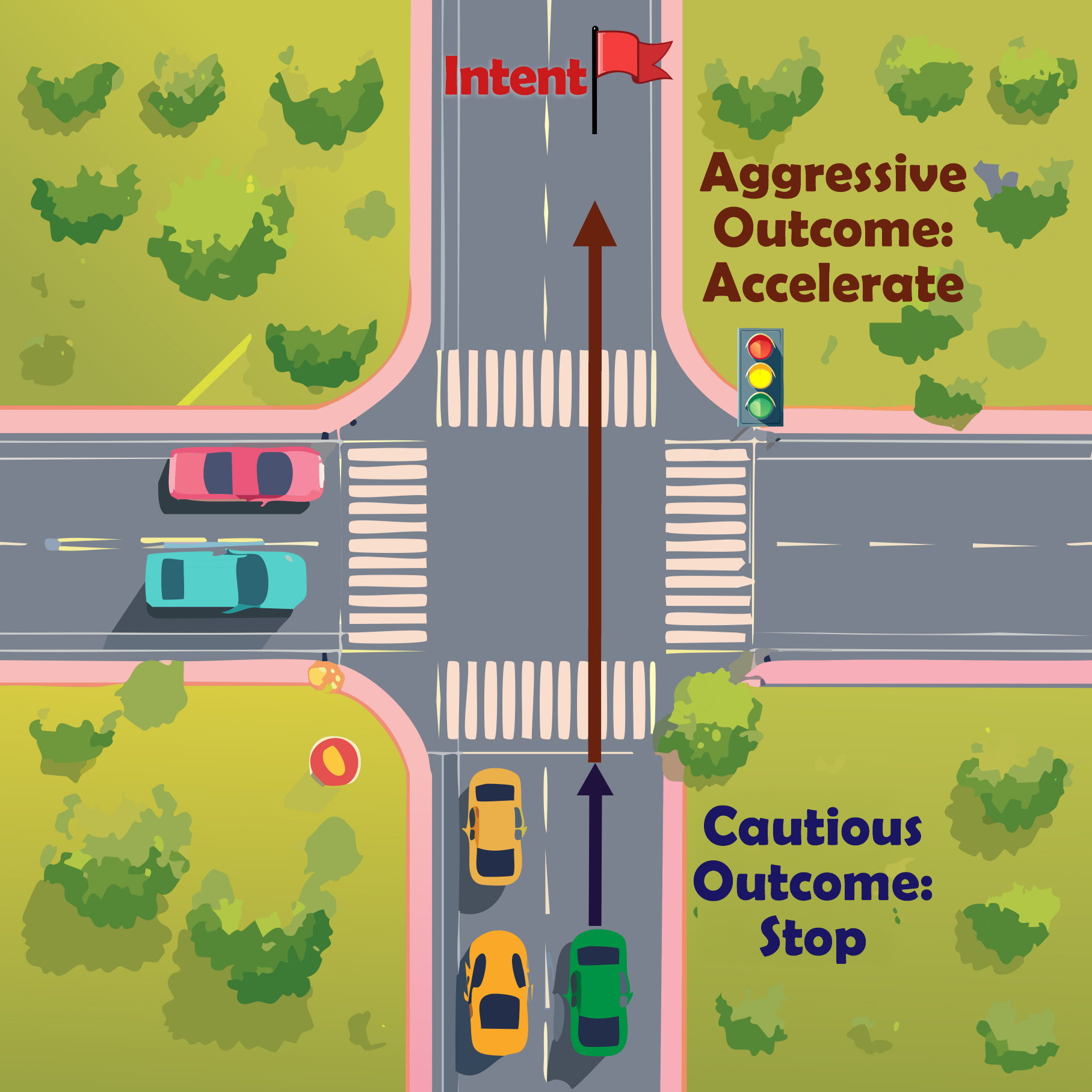}
    \caption{\textbf{Motivating Example: yellow-light scenario.} In this specific but common scenario, a vehicle approaches the intersection such that the yellow light would turn red as the vehicle enters the intersection. This scenario poses a decision to the human driver: slow down to reduce jerk when stopping, or accelerate to cross the intersection before the light turns red. Given that the intention, context, and history of the driver is the same in both outcomes, we hypothesize that multi-modal outcomes here are based on driving style differences. In this example, both the average and the final difference between the two outcomes is high. Driving style is difficult to quantify and under-explored in context of trajectory forecasting for vehicles. In this paper, we perform preliminary analysis of possible quantifications of driving style and propose an approach to address style imbalances and improve performance on rarer driving styles.}
    \vspace{-1em}
    \label{fig:teaser-graphic}
\end{figure}

% Present a motivating example.
For edge-case scenarios, which are arguably riskier than the general scenarios simply due to the fact that the average driver has less experience in them, we argue that the outcomes become increasingly multi-modal. 
Take, for example, a simple yellow light scenario in which the traffic light will turn from yellow to red immediately after the driver approaches the intersection, as in Fig.~\ref{fig:teaser-graphic}.
As drivers identify that they may fall behind the timing of the yellow light, they are faced with the decision to either 1) accelerate so that they reach the intersection faster or 2) begin to yield earlier to avoid a sudden stop.
In this particular example, we may observe a bimodal distribution of outcomes---those with longer trajectories reaching past the intersection light and those whose trajectory cuts off at the light. This example is illustrated in Figure~\ref{fig:teaser-graphic}.

% State our hypothesis that driving style is an important factor.
We hypothesize that \textit{driving style} becomes an increasingly important variable in distinguishing different outcomes in these scenarios. 
To be specific, we consider the notion of driving style to be independent from both the \textit{intent} and \textit{context} of the driver.
We also assume that the driving style of a driver is constant across time; that is, the driving style cannot change within the same trajectory.
If driving style were to be reliably modeled for trajectory prediction, we may even be able to simulate different outcomes in edge cases, even if they were not directly observed in log replays. 
The prospect of such a model would be extremely impactful in testing autonomous driving policies \textit{without needing risks or accidents to occur in the real world first}.

% Gaps in current benchmarks and approaches
Several challenges lay ahead of modeling driving style explicitly in trajectory prediction, however. 
Firstly, we recognize that there is no standard method of quantifying driving style from trajectory data. Without a metric to quantify driving style, we are unable to both incorporate driving style explicitly in learning and properly evaluate policies across stratified driving styles. 
Secondly, estimating driving style directly from real-world data is challenging due to the infrequent, if not rare, occurrence of edge-case scenarios. Without a stressor forcing drivers to make a decision, there is often no decision at all (driving straight). 
The key contributions of this work include:
\begin{enumerate}
\item Analyses on existing real-world trajectory datasets with respect to different driving style quantification approaches, which would be useful for future work in modeling driving style;
\item A cross-examination of driving style distributions in real-world trajectory datasets with different driving style evaluation approaches from robotics (TDBM)~\cite{cheung2018_tdbm} and traffic psychology literature (MDSI)~\cite{MDSI};
\item A method for integrating driving style into trajectory forecasting to model driving style explicitly as a conditioning variable and to rectify imbalances in generalization with respect to driving style. 
\end{enumerate}

With these analyses and the new trajectory prediction model based on driving styles, we hope to make a step towards demystifying how driving style may be quantified and distributed within benchmarks and models the community is already familiar with.

\section{Related Works}

\subsection{Existing Driving Style Analyses}

In recent years, several studies have focused on extracting driving styles from driving data. Most works \cite{liu2021_trajectory_lane_crossing_style} \cite{xing2020_trajectory_leading_vehicles_style} \cite{choi2021_trajectory_gan_style} \cite{kim2021_trajectory_cvae_style} \cite{wang2023_trajectory_detr_style} \cite{hao2024_trajectory_bayesian_style} \cite{chen2022_label_style_graph_cnn} have defined 2-3 categorical driving styles, which are typically variants of aggressive, moderate, and conservative. Some works have also focused on mappings between various driving behaviors and driving styles. Chandra et al. propose a graph-theoretic machine theory of mind that establishes a mapping between various driving indicators and driving styles, and uses vertex centrality functions and spectral analysis to measure the likelihood and intensity of specific driving styles \cite{chandra2020_theory_of_mind_style}.

Many earlier works used clustering algorithms to distinguish driving styles. For example, Xue et al. use K-means clustering on various analytical collision risk surrogates and then use an SVM to classify driving style given acceleration, relative speed, and relative location \cite{xue2019_style_clustering_kmeans_svm}. Similarly, Hao et al. combined offline K-means clustering on average speed, average acceleration, and average time headway with Bayesian filtering for real-time style identification. Other studies \cite{liu2021_trajectory_lane_crossing_style} \cite{xing2020_trajectory_leading_vehicles_style} have predicted style using unsupervised clustering methods based on the Gaussian Mixture Model (GMM) on various kinematic features, such as velocity, acceleration, and jerk.

Beyond traditional clustering techniques, some works have explored the use of deep neural networks to categorize driving styles. In \cite{choi2021_trajectory_gan_style}, a Recurrence Plot is used to convert braking, acceleration, and steering data into images that are then fed through a CNN network. In \cite{kim2021_trajectory_cvae_style}, a DeepConvLSTM network with an attention layer is used to classify driving style based on CAN data. Chen et al. employ fine-grained GPS trajectory segments to construct multi-view graphs, which are then processed through a Graph CNN to generate latent graph embeddings that are classified using a semi-supervised classifier \cite{chen2022_label_style_graph_cnn}.

Other works have explored modeling driving style for personalized autonomous driving policies~\cite{sumner2024personalizing, schrum2024maveric}. While this work is highly related, the goal is different in that these works focus on matching the driving style or preferred interface of a specific, given human-driven driving policy for personalization. 

\subsection{Mapping Driving Style to Trajectory Observations}

Some recent works have proposed using generative models to incorporate driving styles into trajectory prediction models. For lane crossing scenarios, \cite{liu2021_trajectory_lane_crossing_style} used a deep conditional generative model to predict a lane crossing point and a final destination point given state information and a latent style variable. These points are then used to fit a cubic curve that represents the predicted trajectory. In \cite{choi2021_trajectory_gan_style}, a conditional GAN (CGAN) generates driving trajectories conditioned on the driving style label. Similarly, \cite{kim2021_trajectory_cvae_style} predicts trajectories using a conditional VAE (CVAE) model that takes in the classified driving style and estimated past trajectory as conditional inputs.

Recurrent Neural Networks (RNNs) are frequently used in trajectory prediction as they can model sequential data and capture long-term dependencies. In \cite{hao2024_trajectory_bayesian_style}, an LSTM encoder-decoder predicts trajectories given a driving style, lane-changing intentions, and historical trajectories. Similarly, \cite{xing2020_trajectory_leading_vehicles_style} predicts trajectories using a joint time series modeling method that involves selecting a fully-connected regression networks based on a driving style with a shared LSTM layer.

More recently, works have explored using Transformer encoder-decoder architectures to learn trajectories. In \cite{wang2023_trajectory_detr_style}, encoded scene and agent state is passed through a DETR Transformer, where a driving style decoder and future trajectory decoder jointly predict multi-modal trajectories for surrounding vehicles. Unlike previous works which have predefined driving style categories, in \cite{wong2023_trajectory_multistyle}, a Transformer-based style proposal sub-network generates multiple latent ``styles'' simultaneously, each representing different trajectory end-points for the agent. Then, they use a Transformer-based stylized prediction sub-network to predict trajectories given interaction data and multi-style end-point proposals.

\begin{figure}[t!]
    \centering
    \includegraphics[width=1.0\linewidth]{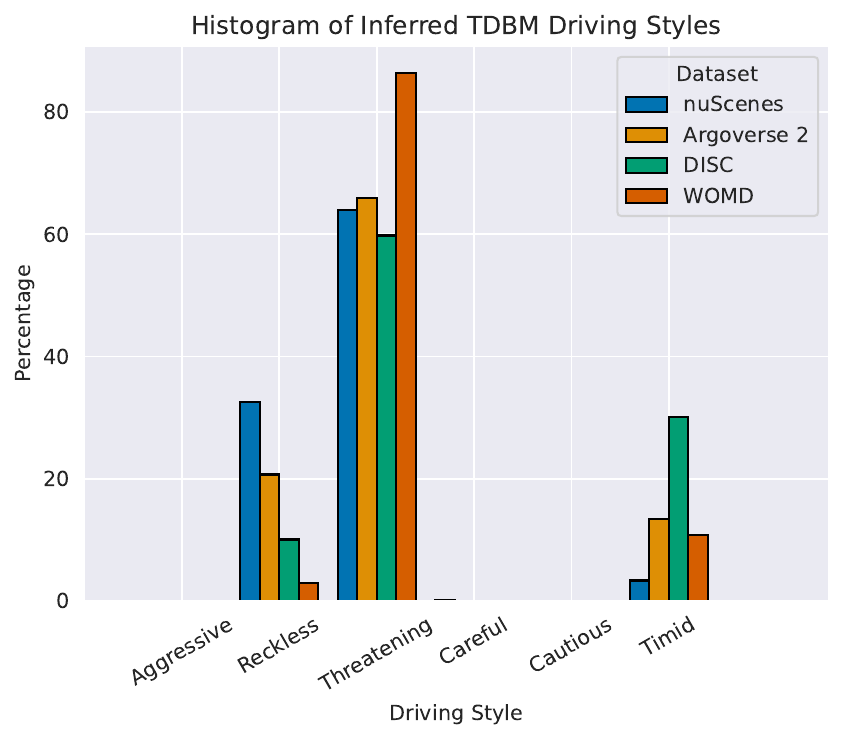}
    \caption{\textbf{Distribution of TDBM Driving Styles in Existing Trajectory Forecasting Evaluation Datasets.} Driving style scoring based on the Trajectory to Driver Behavior Mapping (TDBM)~\cite{cheung2018_tdbm} provides a feature mapping from kinematic properties of trajectories to a driving style classification validated by a user perception study. TDBM has six total driving style classes (from most to least aggressive): aggressive, reckless, threatening, careful, cautious, and timid. 
    The driving style distribution was plotted across four different datasets: nuScenes~\cite{nuscenes}, Argoverse 2~\cite{Argoverse2}, DISC~\cite{kumar2025_disc}, and Waymo Open Motion Dataset (WOMD)~\cite{Ettinger_2021_ICCV_waymo}. While nuScenes, Argoverse, and WOMD represent general real-world driving cases, DISC focuses on pre-crash scenarios collected in virtual reality simulated user studies.
    We note that there is \textit{no} presence of ``aggressive", ``cautious", or ``timid" driving in any datasets, as defined by TDBM. \textit{This may suggest that TDBM is not adequate to capture diverse driving styles in dense urban interactions and/or edge cases.}
    }
    \vspace{-1em}
    \label{fig:tdbm-dist-train}
\end{figure}

\section{Background}

Driving style is difficult to quantify due to it being a human construct.
In general, it is a way to describe how a person drives compared to the average driver in the population. 
Previous works like Trajectory to Driver Behavior Mapping (TDBM)~\cite{cheung2018_tdbm} explore the quantification of driving style with user studies and controlled traffic simulation scenarios with varying kinematic properties.

We use their mapping as a method of quantifying driving styles and bridge the driving behavior classification of modern datasets to recent works exploring driving style and pre-crash scenarios. 

For each agent of interest, we extract a set of kinematic observations from the trajectory data. These features capture key aspects of the agent's motion, such as velocity, acceleration, and jerk.
Let $x \in \mathbb{R}^N$ represent the vector of kinematic observations for a trajectory, where $N$  is the number of features. The TDBM framework uses a predefined matrix $B = \mathbb{R}^{m \times n}$, derived from Equation (5) in the original TDBM paper, where $m$ is the number of driving style classes. The score vector $s \in \mathbb{R}^m$, which represents the scores for each driving style class, is computed as the dot product between the trajectory kinematic features and the mapping matrix:
$s=B \cdot x$.

Here, each element $s_i$ in the score vector corresponds to the score for the $i$-th driving style class. For convenience, we reiterate the mapping matrix $B = [b_0, b_1, b_2, b_3, b_4, b_5]^T \cdot x \\$ from Cheung et al. below: 

{\footnotesize 
\begin{equation}
    B = \begin{pmatrix}
        1.63 & 4.04 & -0.46 & -0.82 & 0.88 & -2.58 \\
        1.58 & 3.08 & -0.45 & 0.02 & -0.10 & -1.67 \\
        1.35 & 4.08 & -0.58 & -0.43 & -0.28 & -1.99 \\
        -1.51 & -3.17 & 1.06 & 0.51 & -0.51 & 1.39 \\
        -2.47 & -2.60 & 1.43 & 0.98 & -0.82 & 1.27 \\
        -3.59 & -2.19 & 1.75 & 1.73 & -0.30 & 0.61 \\
    \end{pmatrix}
    \begin{pmatrix}
        s_{center} \\
        v_{nei} \\
        s_{front} \\
        v_{avg} \\
        j_l \\
        1
    \end{pmatrix} \\
\end{equation}
}

\noindent
where $b_0$ corresponds to Aggressive, $b_1$ corresponds to Reckless, $b_2$ corresponds to Threatening, $b_3$ corresponds to Careful, $b_4$ corresponds to Cautious, and $b_5$ corresponds to Timid.
We note that the naming is ordinal, or represent bins of a spectrum, with ``aggressive" and ``timid" being the two extremes at either end. Thus, we note that while ``threatening" may seem to have a severe connotation, in this context, it is the slightly aggressive average case. Likewise, ``careful" is the slightly less aggressive average case. 
% \begin{figure}
%     \centering
%     \includegraphics[width=1.0\linewidth]{figures/driving_style_histogram_evaluation.pdf}
%     \caption{\textbf{Distribution of TDBM Driving Styles in Existing Trajectory Forecasting Evaluation Datasets.} }
%     \label{fig:tdbm-dist-eval}
% \end{figure}

\section{Analyzing Behavior in Trajectory Datasets}

\subsection{Behavioral Trajectory Analysis and Popular Benchmarks}
As mentioned before, TDBM~\cite{cheung2018_tdbm} validates a mapping between observed kinematic properties of trajectories and different driving behaviors, validated by controlled human perception studies. While this work focused on using predicted driving style for an autonomous vehicle planner, we explore its usage with modern benchmarks and models in trajectory forecasting. To evaluate this, we applied TDBM to the Argoverse 2 dataset, focusing on its ability to capture various driving styles and its alignment with ground truth annotations. Our preliminary analyses suggests that while TDBM effectively identifies high-level behavioral patterns, certain nuances of modern datasets—such as edge-case scenarios and dense urban interactions—pose challenges. Detailed quantitative results are presented in subsequent sections.

We visualized the distribution of driving styles within the nuScenes~\cite{nuscenes}, Argoverse 2~\cite{Argoverse2}, and the DISC~\cite{kumar2025_disc} datasets using TDBM classifications in Figure~\ref{fig:tdbm-dist-train}. The results indicate that existing datasets, under the TDBM classification, are only representative of three of the six total driving behavior classes, suggesting that either current driving datasets are lacking in representation of fringe behavior types or that TDBM is not well-suited for analysis of general trajectory samples. 
In addition to this, we notice that driving style follows a similar pattern across different datasets, even between the real world datasets (nuScenes and Argoverse). 

\begin{figure}
    \centering
    \includegraphics[width=1.0\linewidth]{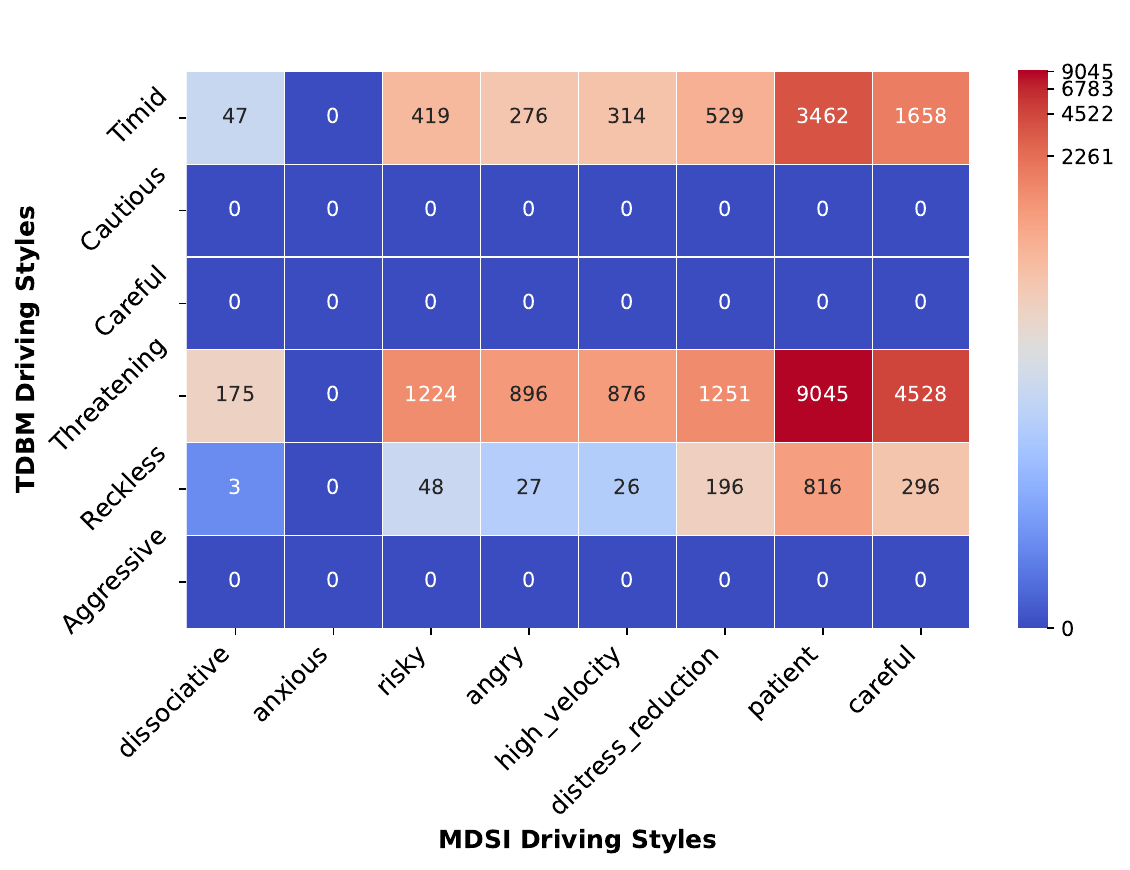}
    \caption{\textbf{Heatmap illustrating the comparison between MDSI-derived self-reported driving styles (X-axis) and TDBM-derived trajectory-based driving styles (Y-axis) within the DISC dataset.} The intensity of each cell represents the frequency of participants exhibiting a given MDSI-TDBM driving style pair. Discrepancies between self-reported and observed behaviors highlight the influence of accident scenarios on driving trajectories and kinematics.}
    \vspace{-1em}
    \label{fig:tdbm-traverse}
\end{figure}

Figure ~\ref{fig:tdbm-traverse} presents a comparative analysis of driving styles derived from the Trajectory-Based Driving Behavior Model (TDBM) and the Multidimensional Driving Style Inventory (MDSI)~\cite{MDSI} within the DISC dataset~\cite{kumar2025_disc}. The DISC dataset is curated through a structured procedure conducted in a virtual reality (VR) simulated environment known as Traverse~\cite{Traverse}. Participants initially complete the MDSI questionnaire, which comprises approximately 50 questions that assess various aspects of driving behavior and decision making, facilitating the classification of their self-reported driving style.

Following the questionnaire, participants engage in controlled accident scenarios within the VR environment, during which their driving trajectories are recorded. These recorded trajectories are subsequently analyzed to infer driving styles using the TDBM approach, enabling a direct comparison with the MDSI-derived classifications. Notably, the figure highlights discrepancies between self-reported and trajectory-based driving styles. Specifically, while certain participants self-identified as patient in the MDSI assessment, their driving behavior in accident scenarios was classified as threatening, timid, or reckless. This discrepancy may be attributed to the distortions in trajectory and kinematic data induced by the nature of accident scenarios, which influence the observed driving behaviors.

\begin{figure*}[t!]
    \centering
    \scalebox{0.8}{
    \includegraphics[width=0.6\linewidth]{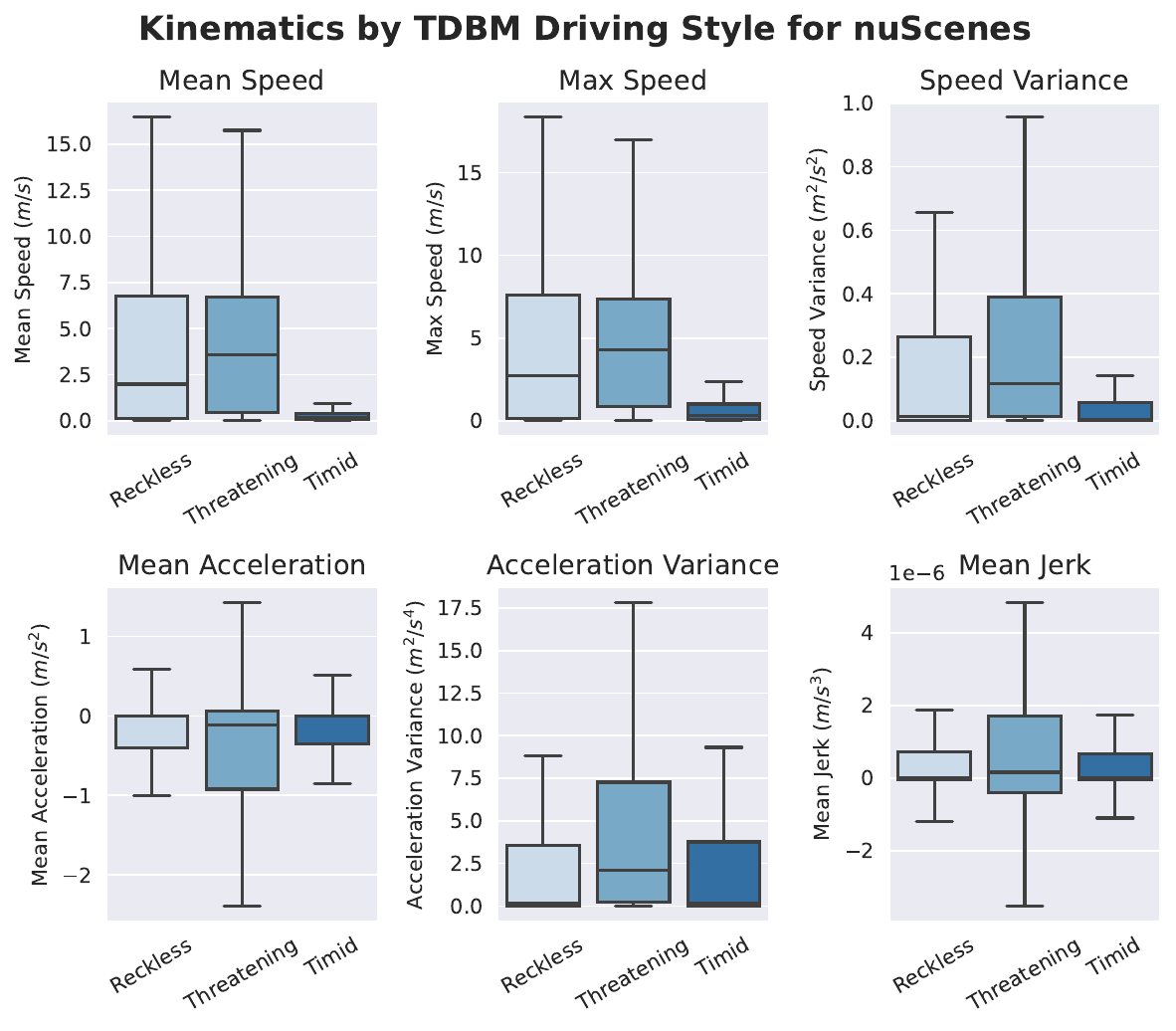}
    \includegraphics[width=0.6\linewidth]{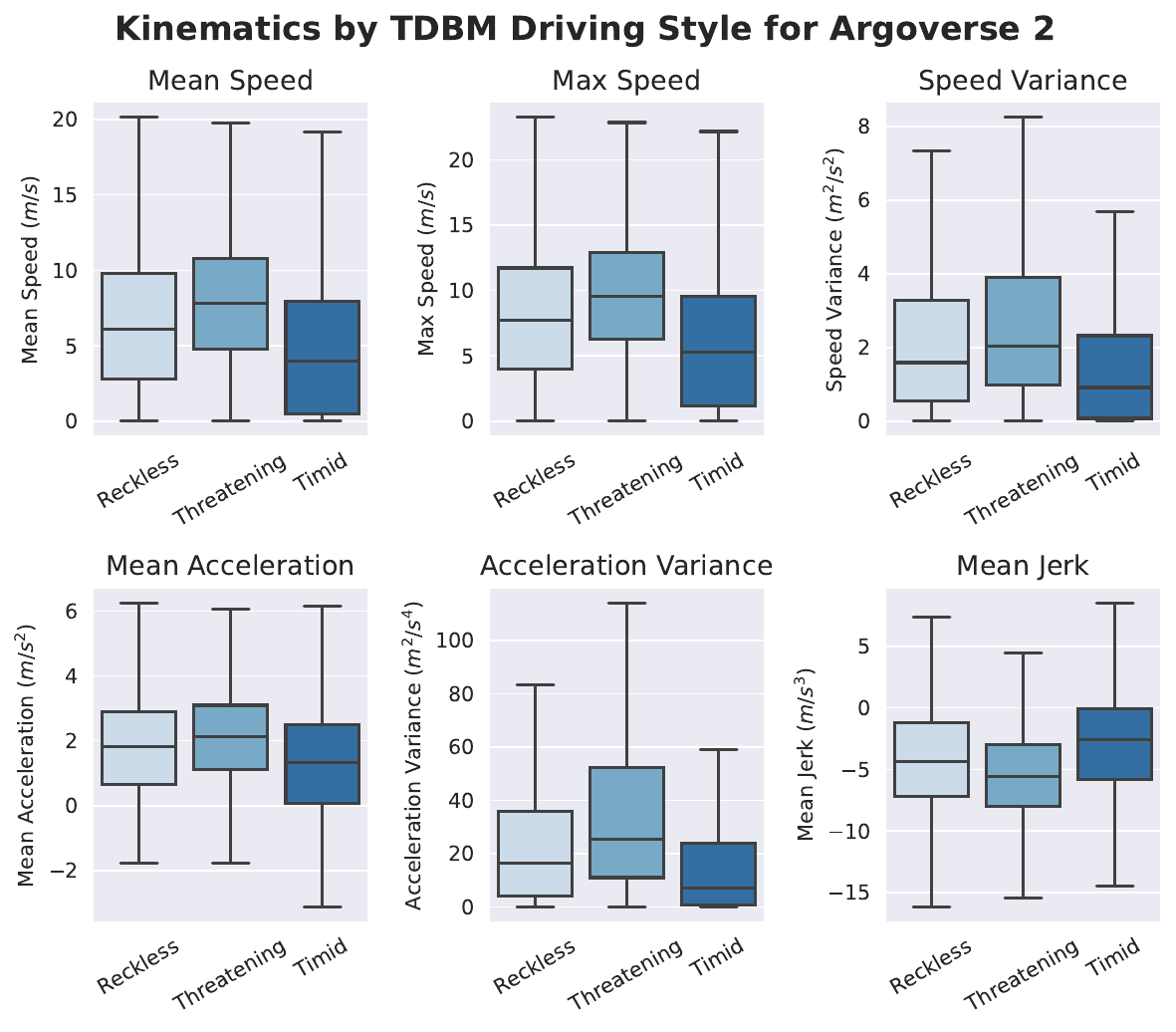}
    }
    \caption{\textbf{Boxplots for Second-Order Kinematics by TDBM Driving Style in the nuScenes and Argoverse 2 datasets.} Different TDBM driving styles have statistically significant differences in kinematics. More aggressive driving styles (reckless/threatening) typically exhibit greater speeds, accelerations, and jerks compared to timid driving. ``Careful" driving style was omitted due to an inadequate number of samples for a statistically significant analysis.}
    \vspace{-1em}
    \label{fig:tdbm-kinematics-boxplots}
\end{figure*}

To further characterize driving styles, we computed second-order kinematic statistics (e.g., speed, acceleration, and jerk) for each TDBM-identified style and graphed the distributions in Figure~\ref{fig:tdbm-kinematics-boxplots}. For the nuScenes dataset, we observe that more aggressive driving exhibits larger mean/max speeds, accelerations, and jerks compared to timid driving, along with greater variance in both speed and acceleration. Interestingly, threatening driving (slightly more aggressive than normal driving) has the widest distribution in mean acceleration, jerk, and speed/acceleration variance. A potential explanation is that there are a variety of behaviors that can be classified as threatening, whereas reckless and timid driving are more uniform and easily identifiable. The Argoverse 2 dataset reveals similar general trends, but compared to nuScenes, the timid driving style has a wider distribution in each metric, demonstrating that different datasets may capture different style trends.

% To validate the relationship between TDBM classifications and second-order kinematics, we compared them against second-order k-means clustering results. The correspondence between these methods was quantified using [placeholder for metric], yielding [placeholder for findings]. This comparison underscores the strengths and potential limitations of TDBM in capturing finer-grained driving behaviors.

% \textbf{Visualization of TDBM Trajectories.} To complement quantitative analyses, we generated visualizations of driving groups identified by TDBM. These visualizations highlight distinct behavioral patterns, such as [placeholder for examples]. Figure \ref{fig:tdbm-kinematics-boxplots} illustrates representative trajectories for each group, providing qualitative evidence of the model's ability to discern meaningful clusters.

\begin{figure*}[t!]
    \centering
    \begin{minipage}[b]{0.32\linewidth}
        \centering
        \includegraphics[width=\linewidth]{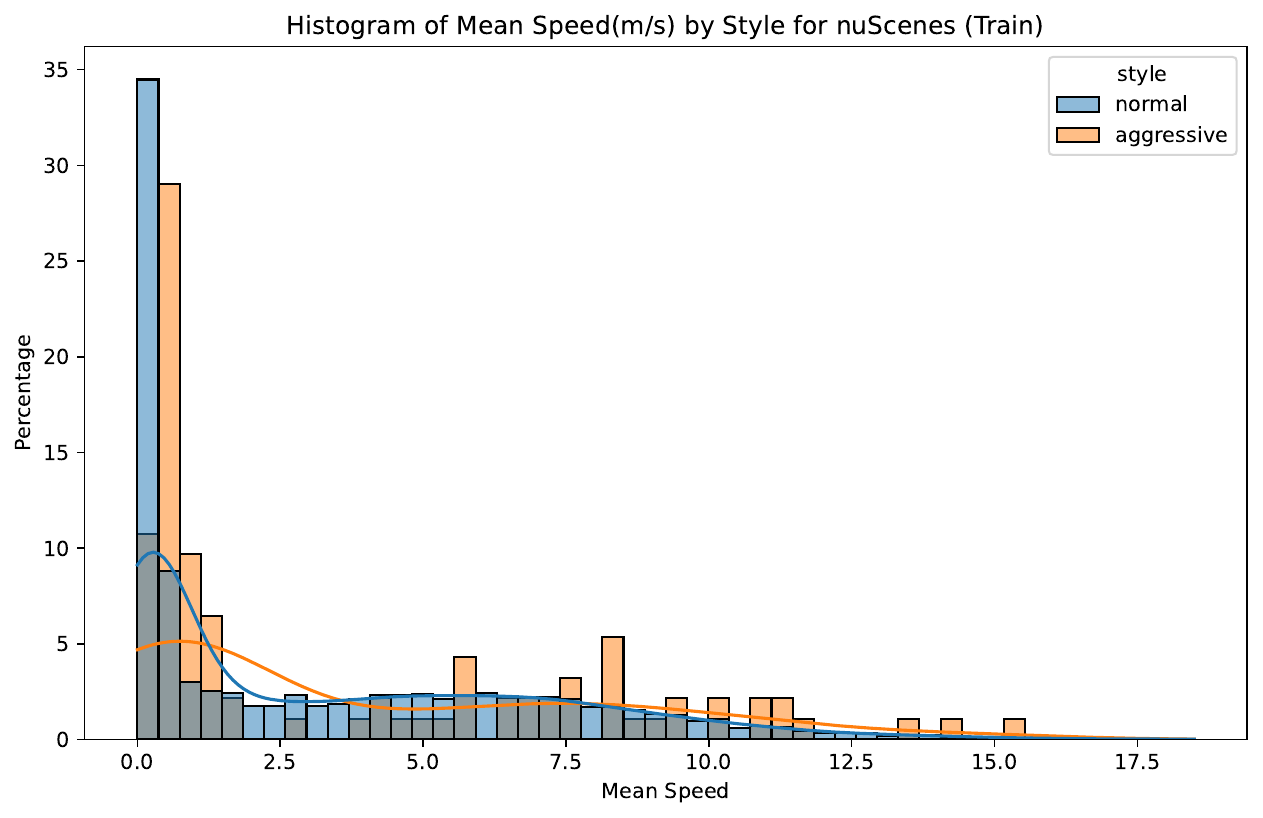}
    \end{minipage}
    \hfill
    \begin{minipage}[b]{0.32\linewidth}
        \centering
        \includegraphics[width=\linewidth]{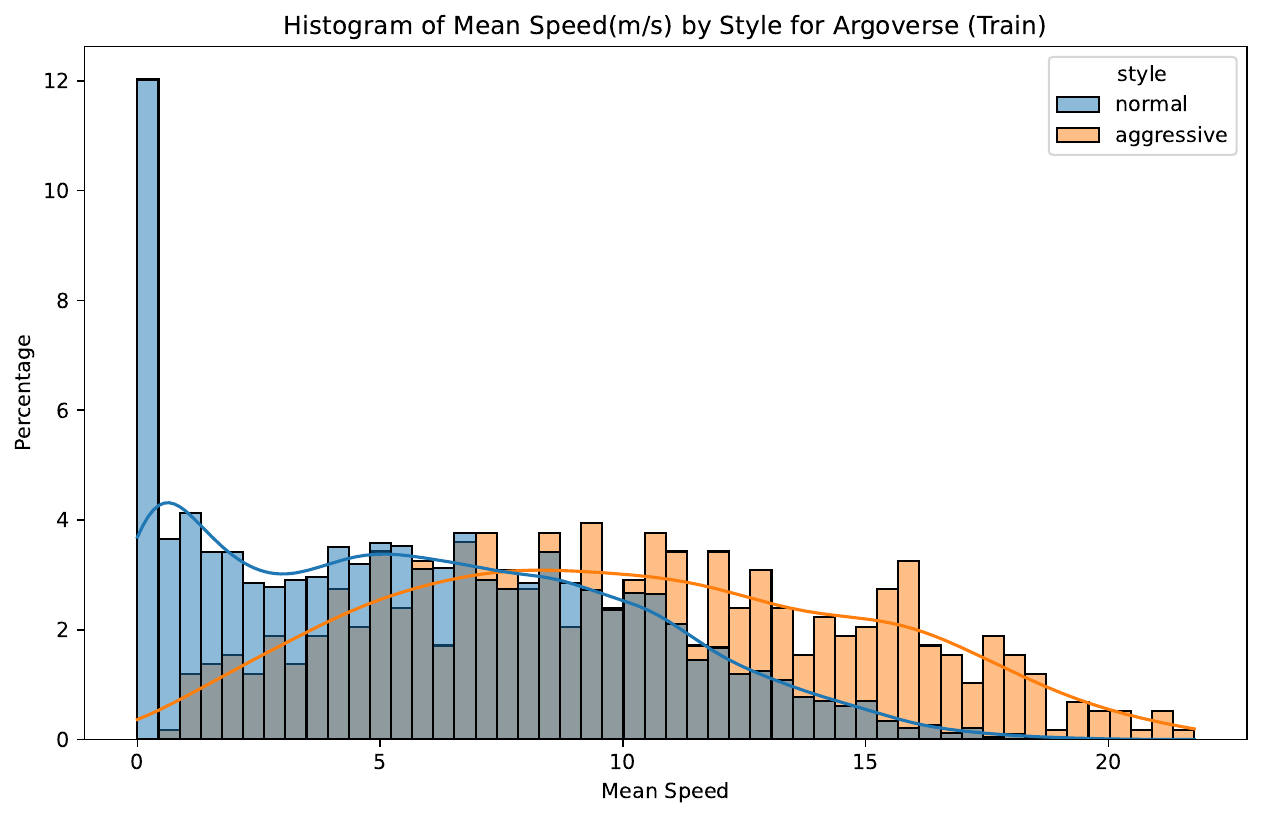}
    \end{minipage}
    \hfill
    \begin{minipage}[b]{0.32\linewidth}
        \centering
        \includegraphics[width=\linewidth]{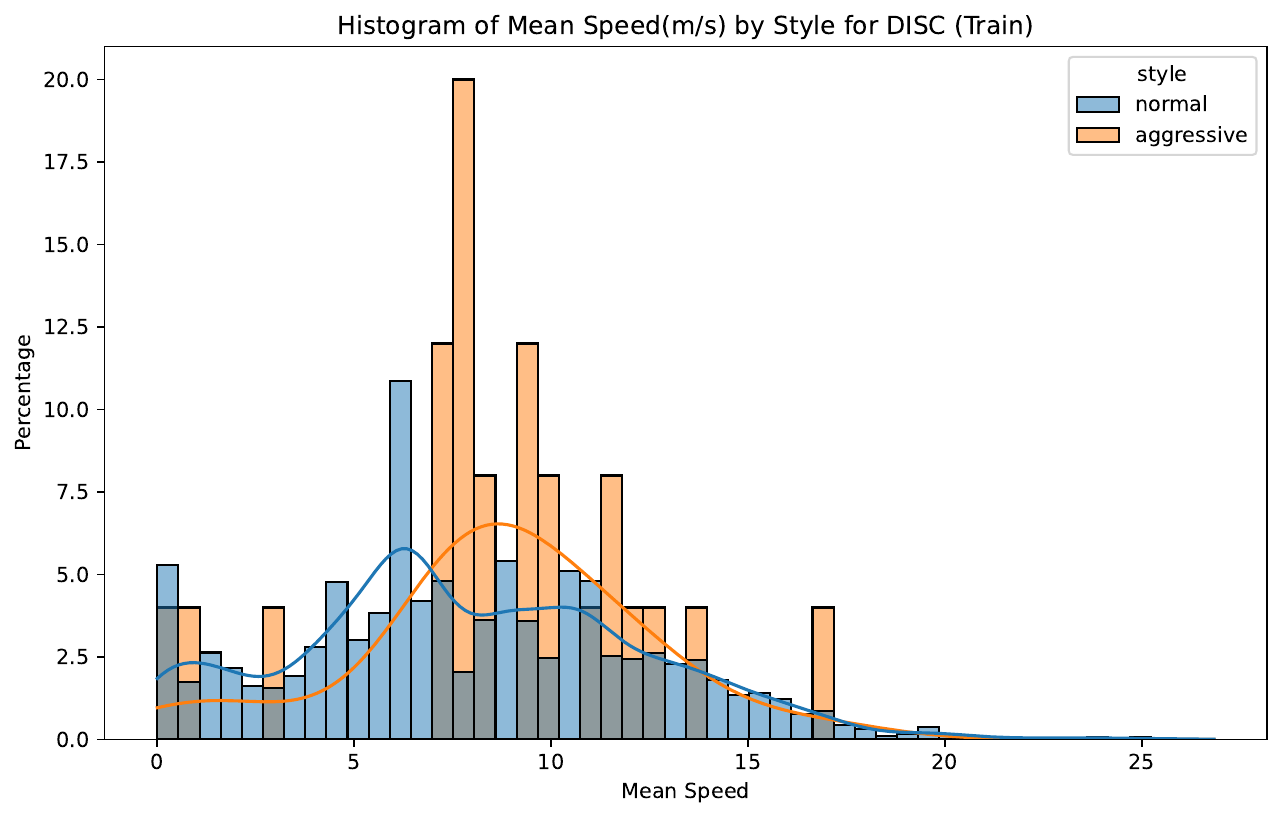}
    \end{minipage}
    
    \vspace{0.5em} % Adjust vertical space
    
    \begin{minipage}[b]{0.32\linewidth}
        \centering
        \includegraphics[width=\linewidth]{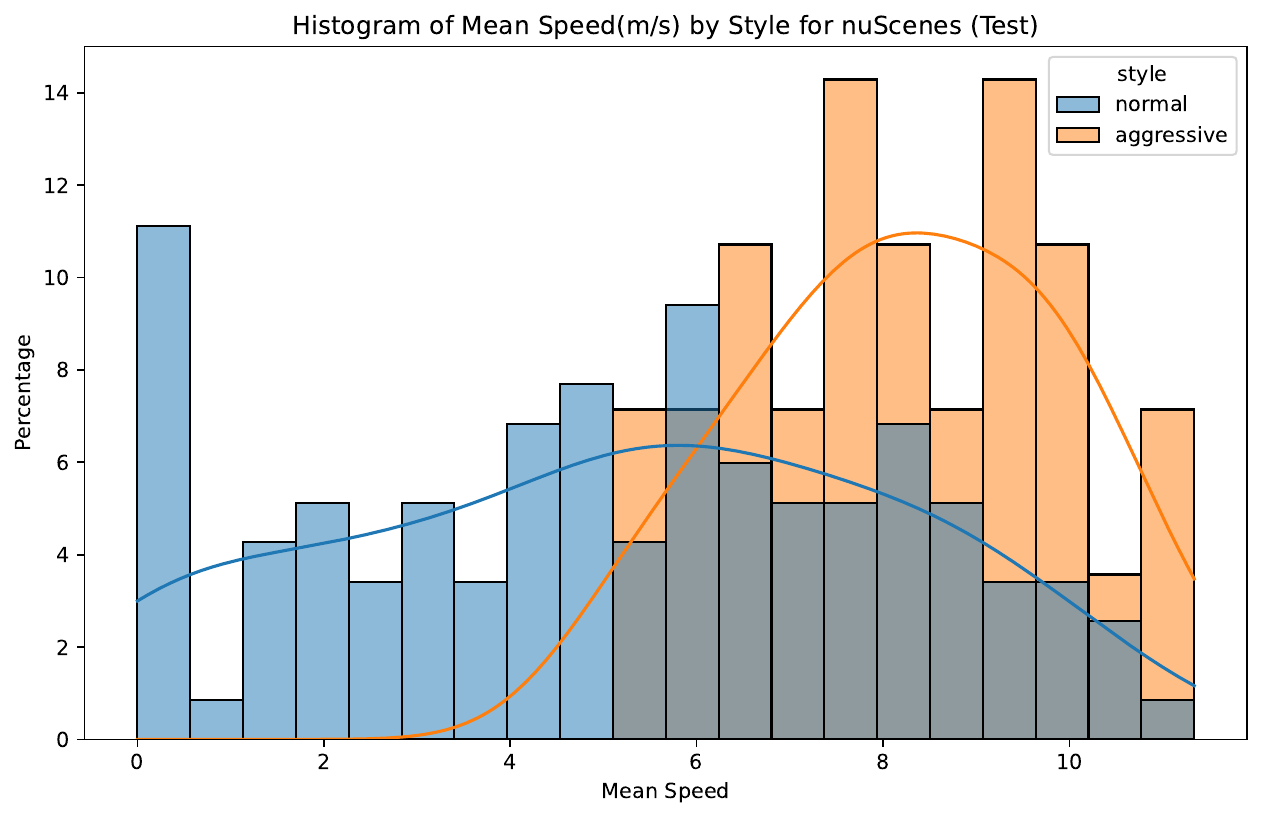}
    \end{minipage}
    \hfill
    \begin{minipage}[b]{0.32\linewidth}
        \centering
        \includegraphics[width=\linewidth]{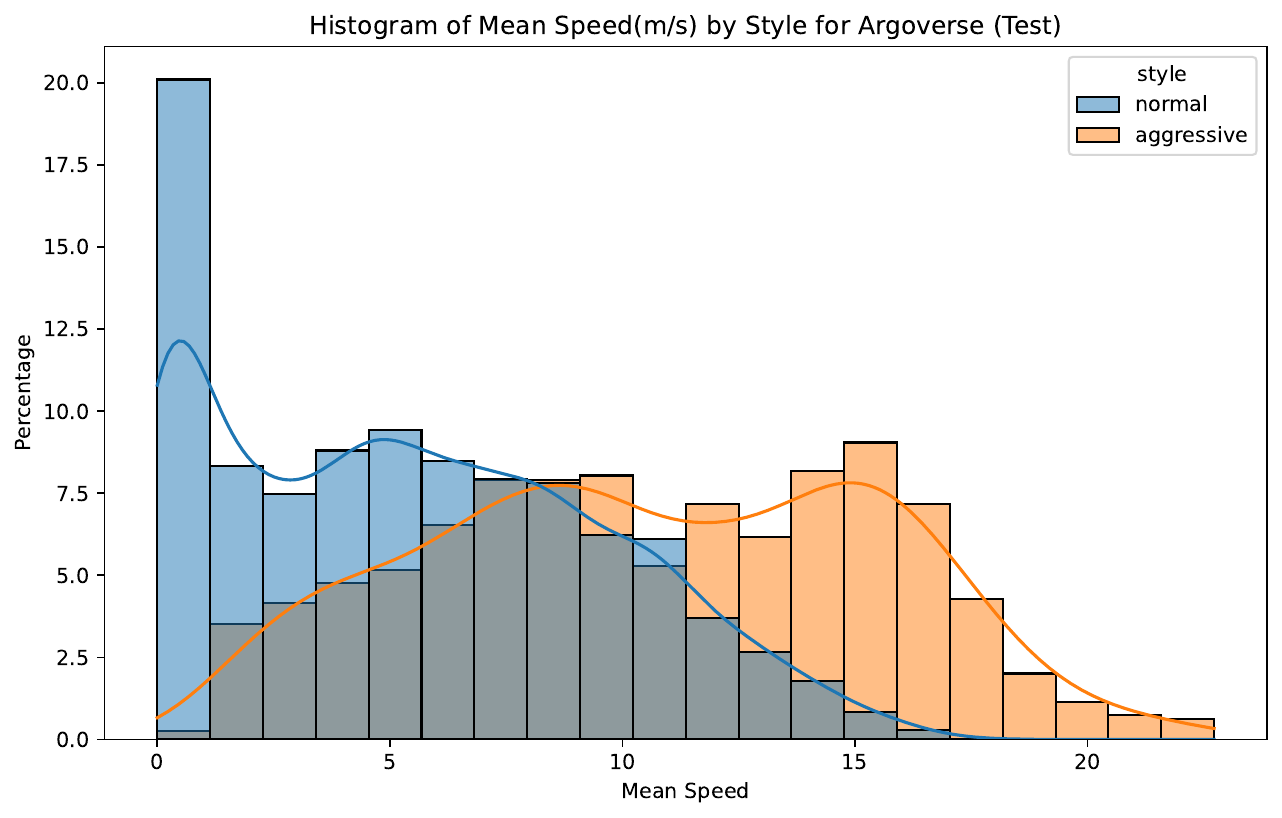}
    \end{minipage}
    \hfill
    \begin{minipage}[b]{0.32\linewidth}
        \centering
        \includegraphics[width=\linewidth]{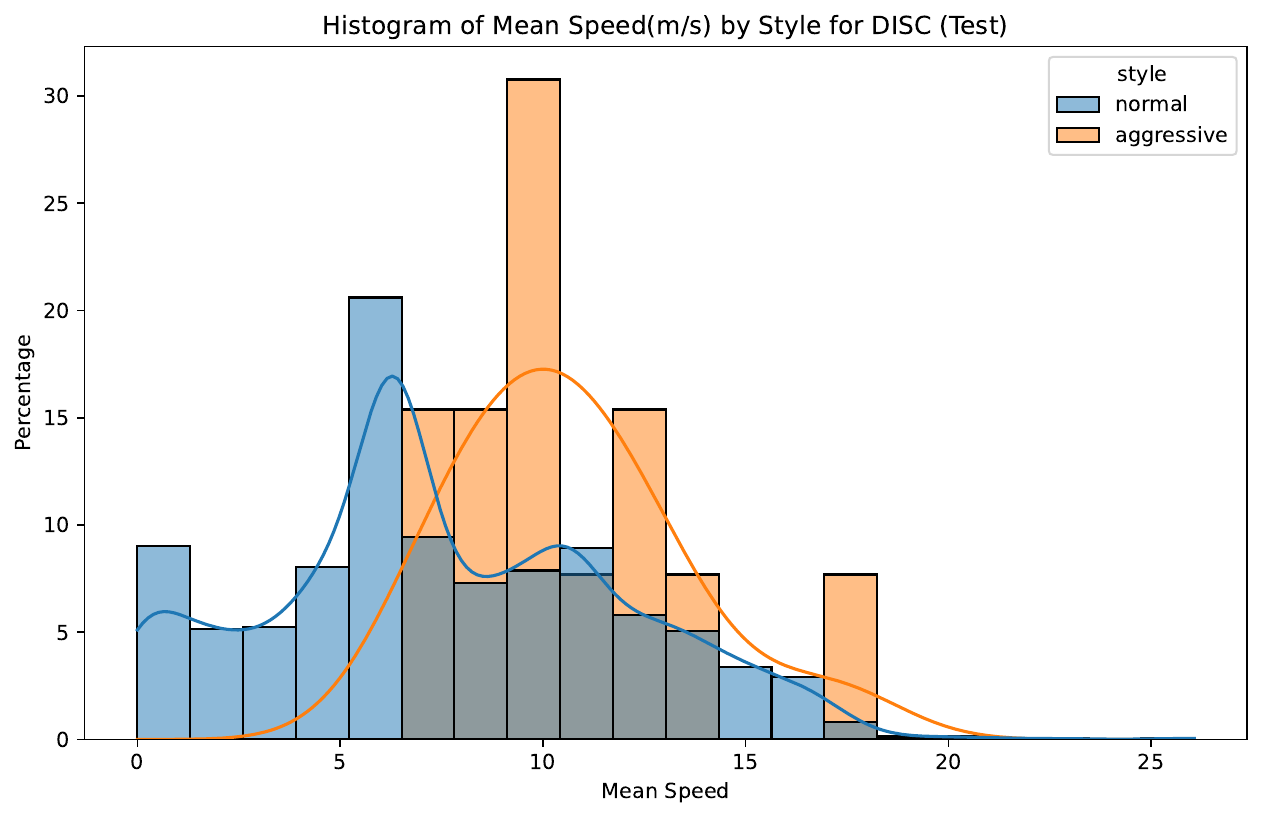}
    \end{minipage}

    \caption{\textbf{Distinct Driving Styles in nuScenes, Argoverse, DISC datasets, with Mean Speed Distributions for Aggressive vs. Normal KDSC Clusters.} Distinct clusters, identified using kinematic features, reveal statistically significant differences in driving styles, demonstrating that embedded dataset information can differentiate aggressive (orange) and normal (blue) driving behaviors. Although these distributions differ markedly across different datasets, they also exhibit meaningful overlap in distribution between training (top) and test (bottom) data, confirming that \textit{our KDSC approach is not naively clustering based on speeds}.  Note:  speed was not used as a direct clustering feature, instead more subtle kinematic cues (i.e. maximum absolute acceleration, variance of acceleration, variance of speed, and $\gamma$) effectively separate different driving styles.}
    \label{fig:mean-speed-distribution}
    \vspace*{-1em}
\end{figure*}

\begin{figure*}[t!]
    \centering
    \begin{subfigure}[b]{0.45\linewidth}
        \centering
        \includegraphics[width=0.95\linewidth]{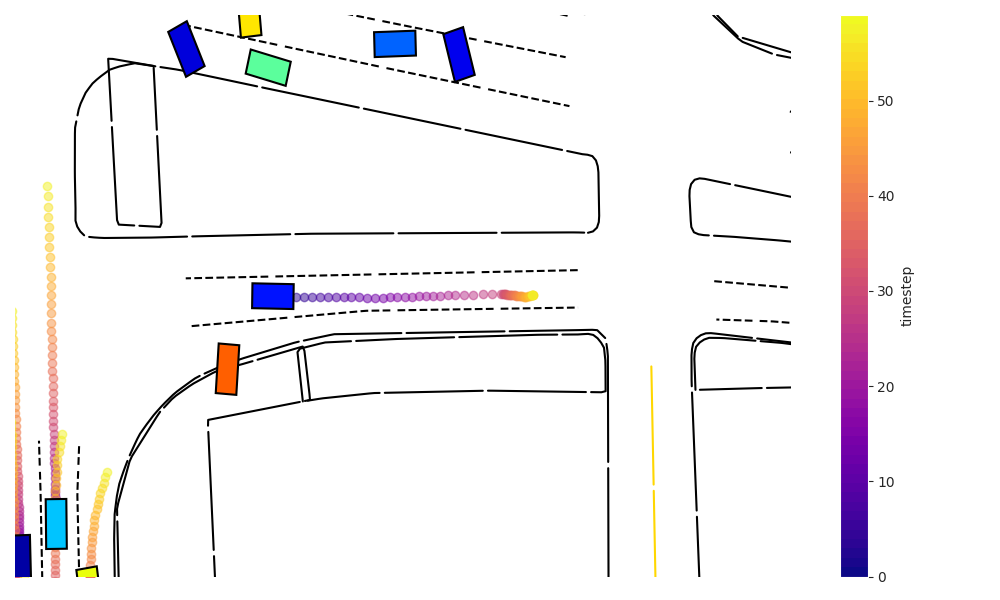}
        \caption{Aggressive Driver}
    \end{subfigure}
    \hfill
    \begin{subfigure}[b]{0.45\linewidth}
        \centering
        \includegraphics[width=0.95\linewidth]{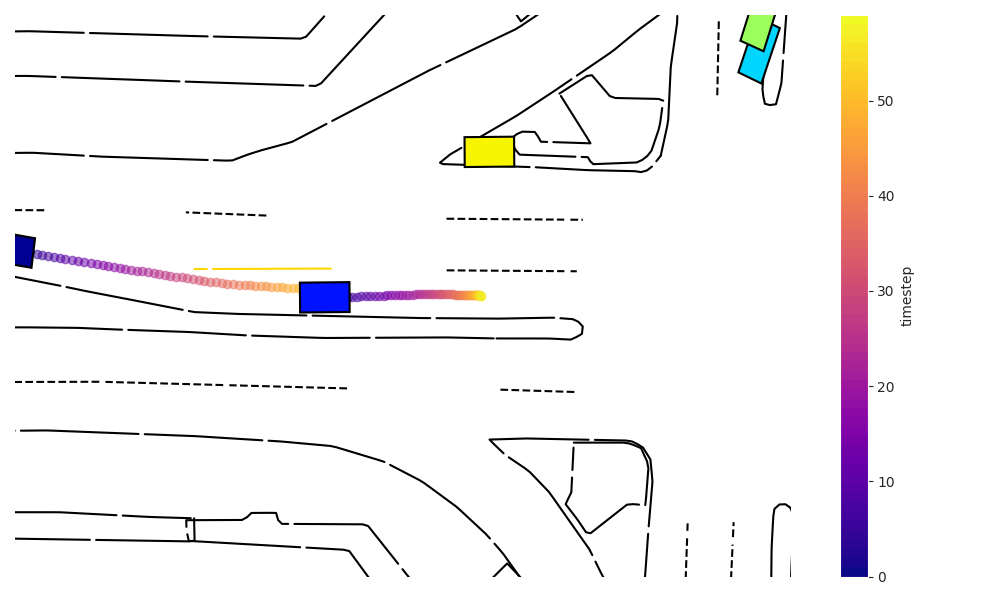}
        \caption{Normal Driver}
    \end{subfigure}

    \vspace{1em}

    % \begin{subfigure}[b]{0.45\linewidth}
    %     \centering
    %     \includegraphics[width=\linewidth]{images/time=7.5.png}
    %     \caption{$t=1.0s$}
    % \end{subfigure}
    % \hfill
    % \begin{subfigure}[b]{0.45\linewidth}
    %     \centering
    %     \includegraphics[width=\linewidth]{images/time=8.0.png}
    %     \caption{$t=1.5s$}
    % \end{subfigure}
    \vspace*{-0.5em}
    \caption{\textbf{Illustrative nuScenes Trajectories for Aggressive vs. Normal Driving Styles from KDSC.} These snapshots from the NuScenes dataset show two different drivers at intersections. The blue box in each subfigure marks the focal vehicle’s initial position, and the colored dots represent its location at $\Delta t = 0.1$ second intervals. In (a), the driver brakes sharply (discontinuity in gradient colors), leading our clustering model to assign the trajectory to the aggressive cluster. In (b), the driver decelerates more smoothly, resulting in its assignment to the normal cluster.}
    \vspace{-1em}
    \label{fig:sample_images}
\end{figure*}

\subsection{Embedded Driving Style Extraction from Kinematic Data
}

While prior work such as TDBM~\cite{cheung2018_tdbm} has successfully mapped driving styles to trajectory data using predefined matrices and user studies, our analysis demonstrates that driving behaviors are inherently encoded within public datasets—even without external annotations. By solely extracting embedded kinematic information, we show that distinct driving styles exist in datasets like nuScenes~\cite{nuscenes}, Argoverse~\cite{Argoverse2}, and DISC~\cite{kumar2025_disc}. We omit results for the Waymo Open Motion dataset due to computational storage limitations. 

To support this claim, we extracted a suite of kinematic features from each trajectory sample, including speed (mean, maximum, and variance), acceleration (mean, maximum absolute, and variance), as well as higher-order dynamics such as jerk (mean and variance) and the dimensionless metric $\gamma$. The inclusion of jerk, which is the derivative of acceleration, and $\gamma$ is particularly crucial because these metrics capture subtle, rapid changes in motion that often indicate aggressive driving maneuvers~\cite{Murphey2009Jerk}.

By leveraging these extracted features, we applied an agglomerative clustering algorithm with $k=2$ to partition the data into two distinct groups. We selected $k=2$ as our analysis confirmed that two clusters provided a good enough separation for following interpretation. Our experiments indicate that using a subset of features including Maximum Absolute Acceleration (max\_abs\_acceleration), Acceleration Variance (var\_acceleration), Speed Variance (var\_speed), and $\gamma$ provided the best separation and consistently yielded well-differentiated clusters that align with intuitive distinctions in driving behavior and provided the most robust separation. We refer to our approach as Kinematic-based Driving Style Clustering (KDSC). In our experiments, we utilized an agglomerative clustering algorithm to demonstrate the effectiveness of this framework. However, KDSC is designed to be flexible and adaptable; alternative clustering models might prove even more effective under different conditions or when applied to future datasets with varied characteristics.

The clustering results, shown in Figure \ref{fig:mean-speed-distribution}, were compelling. Notably, although speed was not used directly as a clustering feature, analysis of the resulting clusters revealed statistically significant differences in mean\_speed distributions. The cluster associated with aggressive driving showed higher variability and elevated speed metrics, while the cluster corresponding to normal driving maintained a more moderate speed profile. This observation reinforces that the selected kinematic features capture the essential nuances of driving behavior. These findings confirm that embedded kinematic features can delineate distinct driving styles without prior mapping frameworks, e.g. TDBM~\cite{cheung2018_tdbm}.

This analysis underscores an important insight for trajectory prediction: public datasets naturally encode various driving behaviors. Recognizing and integrating these variations into predictive models can improve robustness and accuracy, particularly in challenging or edge-case scenarios. Unlike TDBM~\cite{cheung2018_tdbm}, which applies a predefined mapping, KDSC discovers driver behavior patterns purely from trajectory data, making it more adaptable to new datasets without manual annotations. In summary, our findings confirm that extracting and clustering kinematic features from trajectory datasets provides a viable and effective means of identifying inherent driving styles, thereby complementing existing methodologies and paving the way for more nuanced trajectory forecasting models. To illustrate how our KDSC method identifies aggressive maneuvers, Figure \ref{fig:sample_images} shows snapshots from the nuScenes dataset in which the blue car executes a sudden braking maneuver between $t=1.0s$ and $t=1.5s$, highlighting the high jerk and acceleration variability characteristic of this cluster.”

\begin{figure*}[ht!]
    \centering
    \includegraphics[width=1.0\linewidth]{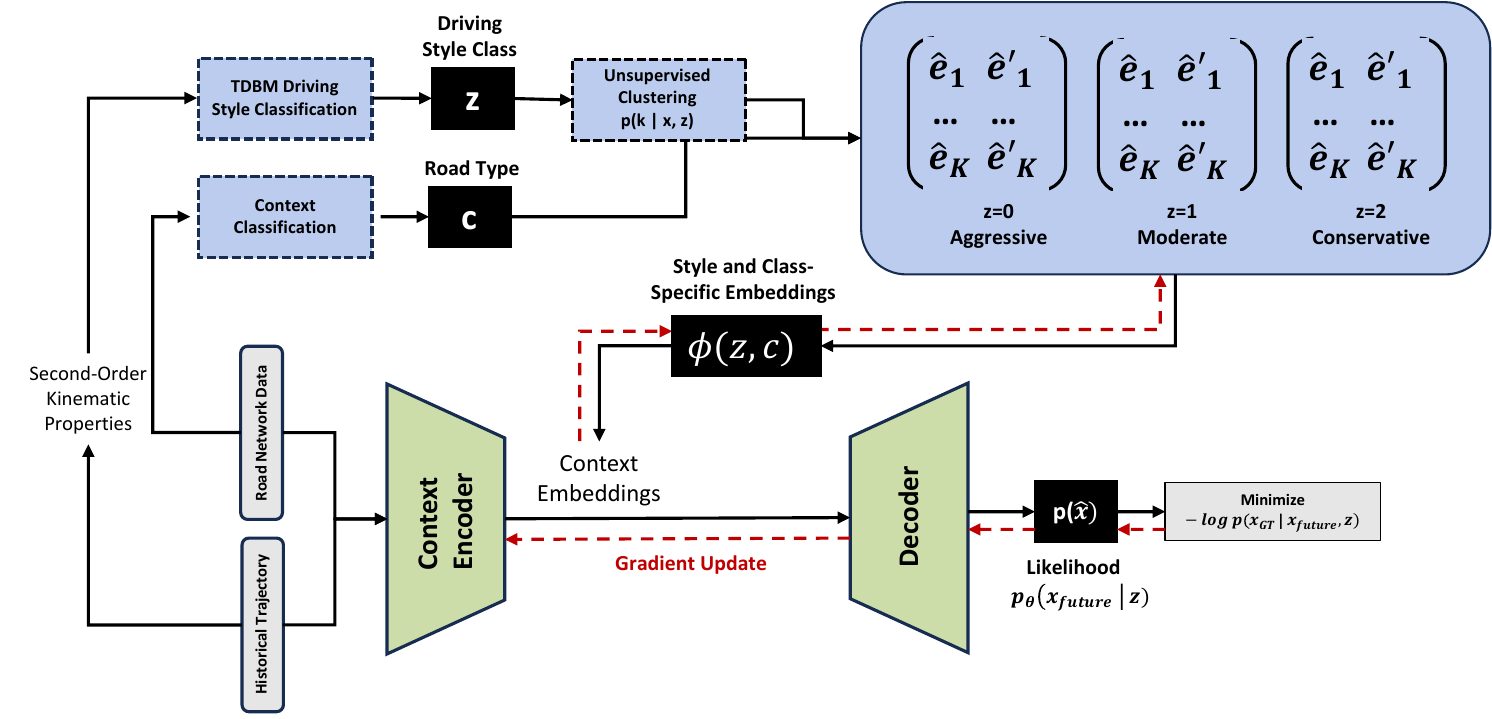}
    \caption{\textbf{Style-Aware Trajectory Prediction with Style-Context Embeddings.} Our approach uses historical agent trajectory information and environment context to determine an indexing function to a library of matrices, factorized by learnable embeddings. After indexing the embeddings corresponding to a particular driving style and road context, they are added to the context embeddings output by the context encoder of the trajectory forecasting model. In our experiments, we use MTR~\cite{shi2022motion} as the backbone model.}
    \vspace*{-1em}
    \label{fig:method_overview}
\end{figure*}

\section{Style-Aware Modeling: Accounting for Driving Style in Trajectory Prediction}
In the previous section, we analyze popular trajectory forecasting datasets to reveal disproportions with respect to driving style, which may negatively influence the generalization of models to fringe behavior types.
Now, we apply the insights from analysis to account for driving style in trajectory prediction, with no additional annotations from what is already included in standard datasets.

Our proposed framework, pictured in Figure~\ref{fig:method_overview}, is built on top of a base encoder-decoder architecture for trajectory prediction. 
Agent trajectory data is first used to construct hand-crafted features to both context classification and driving style classification modules. These modules are not differentiable and are low-dimensional classification functions. 
The driving style classifier assigns a driving style $z \in \{1, ... |Z|\}$ to each agent of interest for prediction. Then, the road network and agent velocity information is used to determine a simple scenario classification. In our case, we only considered two possible context scenarios: highway and non-highway scenarios. 
With these two discrete values assigned to each sample, we then select the appropriate set of matrix embeddings to complement decoder training. 

 % $\phi_z \in \mathbb{R}^{K\times2}$
We initialize a total of $|Z|$ embedding matrices, where each matrix 
$E_z \in \mathbb{R}^{K\times 2}$ is the inner product of two embedding vectors, the row vector $e_k \in \mathbb{R}^{K\times D}$ and the column vector $e_c \in \mathbb{R}^{2 \times D}$.
The driving-style-specific matrix $E_z$ represents embeddings corresponding to a particular intra-style cluster assignment $k$ (row) and one of the two scenarios (columns).

Our approach requires two different steps: 1) Fitting a clustering model hyper-parameterized by K clusters for each driving class $z \in Z$, for a total of $|Z|$ clustering models, then 2) training driving style and context-specific embeddings to complement context embeddings during training. 
While we could use driving style directly in the indexing function for the matrix factorized embeddings, we show in Figure~\ref{fig:tdbm-kinematics-boxplots} that each driving style has different levels of variance. 
Since we expect that increasingly aggressive driving styles will produce more variance in kinematic features of trajectories. 
To account for differences in variance across styles, we model intra-driving-style clusters.

We show trajectory forecasting results for nuScenes across TDBM driving style classes in Table~\ref{tb:model_performance_by_driving_style}. 
In these experiments, we examine whether driving style information benefits generalization over baseline models and performance outcomes between early and late fusion of style information.
We point out two observations: 1) our model not only improves overall metric performance, but also performance on fringe behavior subsets, and 2) our results suggest that early fusion of style information, where style information is provided to the encoder, is slightly more beneficial to generalization than late fusion, where style information is provided to the decoder.

In terms of adaptability, most SOTA architectures such as Wayformer~\cite{nayakanti2022_wayformer} and Motion Transformer~\cite{shi2022motion, shi2023mtr} adopt transformer encoder-decoder architectures. 
% However, we note that our method can be adopted to any base architecture, since the embeddings can be concatenated with any intermediate context features for prediction.

% \begin{table}[t!]
%  \caption{\textbf{Performance of popular models by TDBM driving style on Argoverse.} }
% \label{tb:model_performance_by_driving_style}
% % \vspace{-0.5em}
%   \centering
%   \scalebox{.8}{
%   \begin{tabular}{c|crrrr}
%     \toprule
%    \textbf{Model} & \textbf{Driving Style} & \textbf{brierFDE$\downarrow$} & \textbf{minADE$\downarrow$} & \textbf{minFDE$\downarrow$} & \textbf{MissRate$\downarrow$} \\
%     \midrule
%     \multirow{4}{*}{Timid} & Autobot & 2.431 & 0.861 & 1.791 & 0.310 \\
%     & Wayformer & 2.539 & 0.879 & 1.946 & 0.345 \\
%     & MTR &  2.094 & 0.789 & 1.637 & 0.293 \\
%     & Ours & \\
%     \midrule
%     \multirow{4}{*}{Threatening} & Autobot & 2.672 & 1.041 & 2.003 & 0.342 \\
%     & Wayformer & 2.594 & 0.968 & 1.982 & 0.355 \\
%     & MTR & 2.156 & 0.888 & 1.703 & 0.303 \\
%     & Ours & \\
%     \midrule
%     \multirow{4}{*}{Reckless} & Autobot & 2.649 & 1.019 & 1.992 & 0.359 \\
%     & Wayformer & 2.618 & 0.997 & 2.007 & 0.382 \\
%     & MTR & 2.208 & 0.896 & 1.725 &  0.312 \\
%     & Ours & \\
%     \midrule
%     \multirow{4}{*}{Overall} & Autobot & 2.540 & 0.941 & 1.889 & 0.328 \\
%     & Wayformer & 2.568 & 0.926 & 1.967 & 0.354 \\
%     & MTR &  2.132 & 0.837 & 1.671 & 0.299 \\
%     & Ours & \\

%     \bottomrule
%   \end{tabular}}
% \end{table}

\begin{table}[t!]
 \caption{\textbf{Performance of models by TDBM driving style on nuScenes.} Our results on nuScenes suggest that \textit{early fusion of style information benefits model generalization slightly more than late fusion, especially for ``threatening" behavior (slightly more aggressive than average driving).} }
\label{tb:model_performance_by_driving_style}
% \vspace{-0.5em}
  \centering
  \scalebox{.75}{
  \begin{tabular}{c|crrrr}
    \toprule
   \textbf{Driving Style} & \textbf{Model} & \textbf{brierFDE$\downarrow$} & \textbf{minADE$\downarrow$} & \textbf{minFDE$\downarrow$} & \textbf{MissRate$\downarrow$} \\
    \midrule
    % \multirow{4}{*}{Threatening} & Autobot & 2.261 & 0.602 & 1.388 & 0.500 \\
    % & Wayformer & 3.149 & 0.628 & 2.324 & 1.000\\
    \multirow{3}{*}{Threatening} & MTR & 2.349 & 0.700 & 1.629 & 0.500 \\
    & Ours+EarlyFusion & \textbf{1.977} & \textbf{0.393} & \textbf{1.066} & \textbf{0.000} \\
    & Ours+LateFusion & 3.705 & 1.103 & 2.798 & 0.500 \\
    \midrule
    % \multirow{4}{*}{Reckless} & Autobot & 4.110 & 1.513 & 3.417 & 0.566 \\
    % & Wayformer &  3.229 & 1.003 & 2.611 & 0.448 \\
    \multirow{3}{*}{Reckless} & MTR & 5.209 & 1.686 & 4.541 & 0.671 \\
    & Ours+EarlyFusion & \textbf{4.757} & 1.630 & \textbf{4.090} & 0.559 \\
    & Ours+LateFusion & 4.893 & \textbf{1.566} & 4.158 & \textbf{0.524} \\
    \midrule
    % \multirow{4}{*}{Overall} & Autobot & 4.084 & 1.501 & 3.389 & 0.566 \\
    % & Wayformer & 3.227 & 0.997 & 2.607 & 0.455 \\
    \multirow{3}{*}{Overall} & MTR & 5.170 & 1.673 & 4.501 & 0.669 \\
    & Ours+EarlyFusion & \textbf{4.719} & 1.612 & \textbf{4.048} & 0.552 \\
    & Ours+LateFusion & 4.877 & \textbf{1.559} & 4.139 & \textbf{0.524} \\
    \bottomrule
  \end{tabular}}
\end{table}

\section{Discussion}

One of our objectives is to unveil the distributional properties of popular trajectory forecasting datasets with respect to trajectory forecasting. 
Since most of these datasets are collected from the real world, the driving style classifications reveal to an extent the scarcity of fringe driving behaviors in everyday life.
Thus, any insights from our analysis may not only be useful for benchmarking purposes, but also general modeling considerations for real-world data.

We observe several key takeaways from our analysis:
\begin{enumerate}
    \item Benchmark metric performance drops considerably as the driving behavior increases in aggression; likewise, careful driving is easier to predict despite having less representation than ``threatening". 
    \item Real-world datasets rarely, if not at all, contain fringe behavior samples like ``aggressive" or ``timid", despite these perceived behaviors being recorded in the perception study by TDBM~\cite{cheung2018_tdbm}.
    \item Distributions of both kinematic properties and occurrence of driving styles vary greatly between training and validation sets for nuScenes, as shown in Figure~\ref{fig:mean-speed-distribution}. When we take into account the behavior type inferred from our clustering approach, the performance on validation improves. This suggests that the driving behavior is a common factor between both sets and contributes meaningfully to generalization. Additionally, early fusion of style information may work slightly better than late fusion for transformer-based architectures such as Motion Transformer~\cite{shi2022motion}.
\end{enumerate}

\noindent
\textbf{Practical Considerations with TDBM.} 
The kinematic features associated with TDBM are computed relative to neighboring vehicles. 
However, in some cases, there may be no neighboring vehicles at all. In these cases, we set the relative speeds to zero, corresponding to having no difference from a vehicle moving the same speed as its neighbors. 
This may not be an accurate classification of a particular driving style due to the lack of stressors in the environment or the lack of necessity to make decisions. 
When there are no neighboring vehicles in the environment, the TDBM class defaults to ``threatening". Thus, there is already a tendency for classifications to lean towards ``threatening" by a factor of the number of samples with no neighbors. 
% Keeping this limitation in mind, we note that many samples in nuScenes have no neighbors, while almost no samples have no neighbors in Argoverse. 

% While we omit 
Additional analyses, results, code, and processed dataset releases will be posted on our project website:  \url{https://gamma.umd.edu/traj_style_analysis/}.

\section{Conclusion}

In this paper, we % firstly surveyed methods used in existing trajectory analysis and 
conducted preliminary analysis on quantifying driving styles for trajectory forecasting. 
Using insights from analysis, we proposed a style-aware trajectory prediction model, which takes advantage of style and context-specific embeddings. 
Our trajectory prediction results showed better performance on fringe behavior styles, such as ``reckless". 
Additionally, our insights on quantifying driving style revealed inherent imbalances within trajectory prediction datasets with respect to driving styles, even among the training and validation sets of the same dataset. 

Our work can be further extended. First, driving style is not entirely demystified as described in this work. Driving style is highly complex and high-dimensional---our work only scratches the surface of the beginning to quantify and model driving styles in trajectory forecasting. 
Prior scoring method, TDBM~\cite{cheung2018_tdbm}, has limitations when there are no neighboring vehicles on the road, while clustering is inherently tied to kinematics of a vehicle, where the kinematics may also be not disentangled from other environmental factors. However, through preliminary demonstration, this work shows promise in modeling driving styles explicitly in trajectory forecasting or in driving simulation.
Second, our analysis reveals that different driving styles exhibit distinct kinematic distributions, indicating that models trained solely on general cases might overlook the nuances of less-represented behaviors. Leveraging these unsupervised style annotations as an auxiliary signal could enable targeted balancing strategies that improve performance in critical edge-case scenarios, thereby improving safety of autonomous driving. Furthermore, integrating recent out-of-distribution generalization techniques to address data imbalances may ultimately yield a more robust trajectory forecasting framework~\cite{Sagawa2019DistributionallyRN, Ghaznavi2023AnnotationFreeGR, Liu2021JustTT}.

\bibliography{references}
\bibliographystyle{ieeetr}

\end{document}